\documentclass[10pt,twocolumn,letterpaper]{article}

\usepackage{wacv}              

\usepackage{amssymb}
\usepackage{cite}
\usepackage{amsmath}
\usepackage{graphicx}
\usepackage{float} 
\usepackage{subfloat}
\usepackage{overpic}
\usepackage{wrapfig}
\usepackage{balance}
\usepackage{ulem}
\usepackage{color}
\usepackage{bbding}
\usepackage{bm}
\usepackage{multirow}
\usepackage{mathtools}
\usepackage{booktabs} 

\usepackage[pagebackref,breaklinks,colorlinks]{hyperref}

\usepackage[capitalize]{cleveref}
\crefname{section}{Sec.}{Secs.}
\Crefname{section}{Section}{Sections}
\Crefname{table}{Table}{Tables}
\crefname{table}{Tab.}{Tabs.}

\def\wacvPaperID{663} 
\def\confName{WACV}
\def\confYear{2025}

\begin{document}

\title{MSI-NeRF: Linking Omni-Depth with View Synthesis through \\ Multi-Sphere Image aided Generalizable Neural Radiance Field}

\author{Dongyu Yan
\and
Guanyu Huang
\and
Fengyu Quan
\and
Haoyao Chen
\and
\\
Harbin Institute of Technology (ShenZhen)
}
\maketitle

\begin{abstract}
    Panoramic observation using fisheye cameras is significant in virtual reality (VR) and robot perception.
    However, panoramic images synthesized by traditional methods lack depth information and can only provide three degrees-of-freedom (3DoF) rotation rendering in VR applications.
    To fully preserve and exploit the parallax information within the original fisheye cameras, we introduce MSI-NeRF, which combines deep learning omnidirectional depth estimation and novel view synthesis.
    We construct a multi-sphere image as a cost volume through feature extraction and warping of the input images.
    We further build an implicit radiance field using spatial points and interpolated 3D feature vectors as input, which can simultaneously realize omnidirectional depth estimation and 6DoF view synthesis.
    Leveraging the knowledge from depth estimation task, our method can learn scene appearance by source view supervision only.
    It does not require novel target views and can be trained conveniently on existing panorama depth estimation datasets.
    Our network has the generalization ability to reconstruct unknown scenes efficiently using only four images.
    Experimental results show that our method outperforms existing methods in both depth estimation and novel view synthesis tasks.
\end{abstract}


\section{Introduction}

Omnidirectional imagery is an important tool for computer vision, VR, and robotics.
Compared to normal pin-hole images with limited field-of-view (FoV), omnidirectional images include all viewing angles both horizontally and vertically.
It creates a compact way to store information about an entire scene in just one camera shot.
Traditionally, omnidirectional images are synthesized using multiple fisheye cameras or camera arrays \cite{brown2007automatic, lian2018image}.
Seamless omnidirectional images can be generated by matching and stitching overlapping pixels between adjacent images.
They are then saved in equirectangular projected images using spherical coordinate representation.

\begin{figure}[t]
    \centering
    \includegraphics[width=0.5\textwidth]{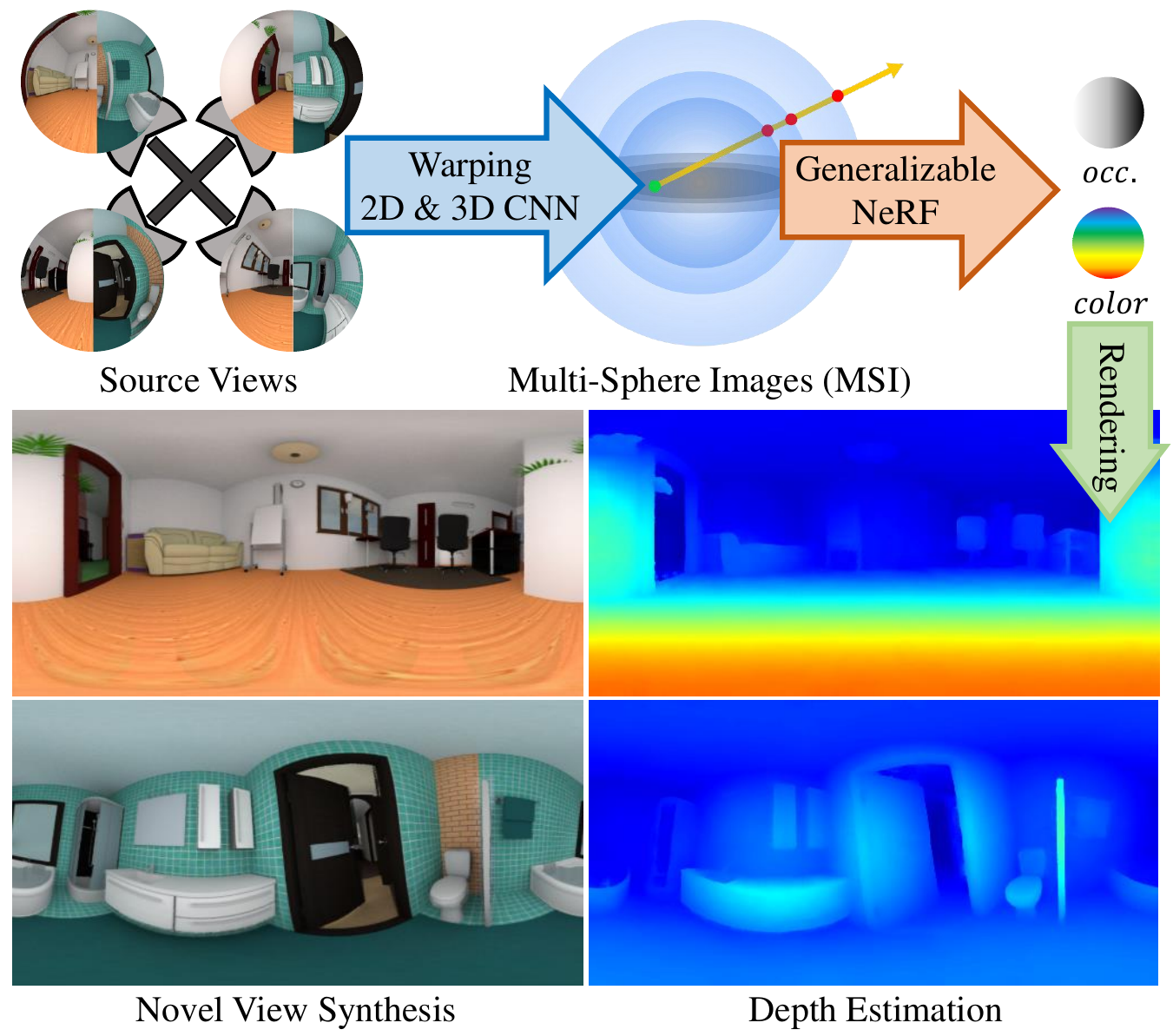}
    \caption{
        Our method uses the images captured from four fisheye cameras arranged in a panoramic configuration as input.
        Aided by a multi-sphere image, a generalizable omnidirectional radiance field can be produced.
        From the radiance field, we can query occupancy and color information of any spatial position and ray direction.
        Then, leveraging the volume rendering formula, novel view synthesis, and depth estimation can be accomplished.
    }
    \label{fig:banner}
\end{figure}

Although the abovementioned method can synthesize omnidirectional images with pleasant visual quality, more is needed.
On the one hand, omnidirectional images assume that the rays corresponding to all pixels have the same origin.
Since it is impossible to place the optical centers of multiple cameras at the same point, there will always be errors in the stitching method, especially in large parallax scenarios.
On the other hand, stitching and blending images from cameras at different positions will cause the originally existing disparity information to be eliminated.
As a result, the final produced image loses its depth information and can only be rendered in 3DoF, limiting its practical usage in the scope of robotics and virtual reality.

With the development of computer vision and deep learning, methods have tried to solve such problems.
OmniMVS \cite{won2019omnimvs} extends the traditional multi-view stereo (MVS) method to panoramic views.
It uses deep learning methods to match features between different cameras and returns the panoramic depth map.
However, it does not consider occluded regions and is insufficient to support 6DoF rendering.
OmniNeRF \cite{abs-2106-10859} proposes a method for 6DoF rendering panoramically.
However, since it requires processed omnidirectional images and additional depth as input, it can not be implemented end-to-end.
The need for per-scene optimization also limits its application.
MVSNeRF \cite{chen2021mvsnerf} achieves free view rendering with generalizable network.
Nevertheless, it only works for object-level or front-facing scene reconstruction.
No research has achieved the reconstruction of panoramic scenes using a single shot.

This paper aims to generate an omnidirectional representation that preserves the 3D information in the input multi-view fisheye images.
We can obtain omnidirectional depth information from such representation and perform 6DoF perspective synthesis.
As input, we collect images from four surrounding fisheye cameras, and follow the general MVS pipeline to build a multi-sphere image (MSI) at predefined depth layers as a cost volume.
A NeRF \cite{mildenhall2021nerf} is then constructed to represent the scene implicitly.
It takes the interpolated features from the cost volume as an additional input to better utilize the learned geometry and texture.
Traditional novel view synthesis task usually needs datasets containing both source and target view images for training.
But mainstream panoramic datasets only contain depth data and rarely include target views.
Our method utilizes input images to introduce color supervision into the network, enabling generalizable NeRF training using only depth ground truth.
The outline of our method is shown in Fig. \ref{fig:banner}.

With our approach, we can efficiently render omnidirectional depth maps and novel view images.
Vision-based panoramic depth has the advantages of lower cost, higher resolution, and larger FoV compared with LiDAR.
Such depth data plays an important role in autonomous driving, scene reconstruction and robotics application.
For the field of virtual reality, being able to render 6DoF novel views eliminates the VR sickness caused by 3DoF rendering.
It also proposes a new way of spatial video acquisition and editing, and can perform video de-shaking for both rotation and translation.

In summary, our main contributions are as follows:
\begin{itemize}
    \item[$\bullet$] We propose a deep learning method to synthesize an omnidirectional radiance field from only four fisheye inputs.
    The traditional 2D panorama output is expanded into 3D, and the parallax information inside the original images is preserved.
    \item[$\bullet$] We combine the depth estimation task with the novel view synthesis task.
    By leveraging the inductive bias from MSI representation, multi-task network training can be achieved only using the most common depth data supervision.
    \item[$\bullet$] Our trained network can generalize across scenes by pre-training on synthetic datasets.
    In experiments, our method performs well on both tasks and achieves state-of-the-art results.
    It can be used in various VR and robotics applications to eliminate VR sickness, enable panoramic video editing, and 3D reconstruction.

\end{itemize}

\section{Related Works}

\subsection{Omnidirectional Depth Estimation}

Depth estimation using multi-view images has long been a concern in computer vision \cite{won2019sweepnet, won2019omnimvs, xie2023omnividar} and robotics \cite{gao2017dual, gao2020autonomous, xu2022omni, liu2024omninxt}.
By stereo-matching colors or features from different viewing angles, depth information can be obtained through the MVS methods \cite{chang2018pyramid, yao2018mvsnet}.

Unlike the general MVS algorithms, which can only generate depth maps of a specific target perspective, omnidirectional depth estimation aims to obtain panoramic depth using cameras surrounding all $360^\circ$.
It usually requires multiple large FoV fisheye cameras \cite{won2019sweepnet, won2019omnimvs, meuleman2021real, hane2014real, xie2023omnividar, chen2023unsupervised} or spherical camera array \cite{zuckerberg2016360, broxton2019low} as input.
SweepNet \cite{won2019sweepnet} modifies the Plane Sweep method \cite{hane2014real} into spherical coordinates and uses Semi-Global Matching (SGM) \cite{hirschmuller2007stereo} to obtain a depth map.
Its following work, OmniMVS \cite{won2019omnimvs}, also uses the quad fisheye array but improves the matching module to a 3D convolution network.
Sphere-Sweeping-Stereo \cite{meuleman2021real} proposes to use adaptive spherical matching and cost aggregation to accelerate inference while maintaining a fine-grained depth map quality.
OmniVidar \cite{xie2023omnividar} improves the traditional fisheye camera model and proposes an attention-based recurrent network structure for high-quality depth estimation.

The above omnidirectional depth estimation methods can, to some extent, solve the problems of parallax information utilization and 3D scene perception.
However, free view roaming and rendering occluded parts can still not be achieved.

\subsection{Omnidirectional Novel View Synthesis}

Novel view synthesis means to continuously interpolate new views given a limited set of source views.
NeRF \cite{mildenhall2021nerf} and multi-plane image (MPI) \cite{zhou2018stereo, li2021mine} methods perform view synthesis on object level and forward facing scenes by constructing implicit or explicit geometric proxy.
Although high quality novel view images can be generated, the bounded scene region still limits their practical usage.

In order to extend novel view synthesis to unbounded scenes, methods using full space mapping \cite{barron2022mip, zhang2020nerf++, wang2023f2, choi2023balanced, barron2023zip} and omnidirectional input \cite{abs-2106-10859, attal2020matryodshka, kulkarni2023360fusionnerf, gu2023vanishing, wang2023perf, hara2022enhancement, muhlhausen2023immersive} have been developed.
Mip-NeRF 360 \cite{barron2022mip} enhances Mip-NeRF \cite{barron2021mip} with scene parameterization to realize $360^\circ$ scene representation.
But it still needs densely captured views for high quality rendering.
Methods taking omnidirectional images alleviates such problem.
MatryODShka \cite{attal2020matryodshka} takes omnidirectional stereo (ODS) images as input and uses neural network to blend them into MSI for novel view synthesis.
OmniNeRF \cite{abs-2106-10859} first synthesizes multi-view images from color and depth panorama by point projection.
It then trains a scene specific NeRF using the two-stage method, which lacks practicality.

Due to that multi-view datasets containing panoramic images are difficult to obtain, methods usually use single view with depth auxiliary for nerf training.
However, these methods usually lead to view penetration and artifacts.

\begin{figure*}[t]
    \centering
    \includegraphics[width=0.9\textwidth]{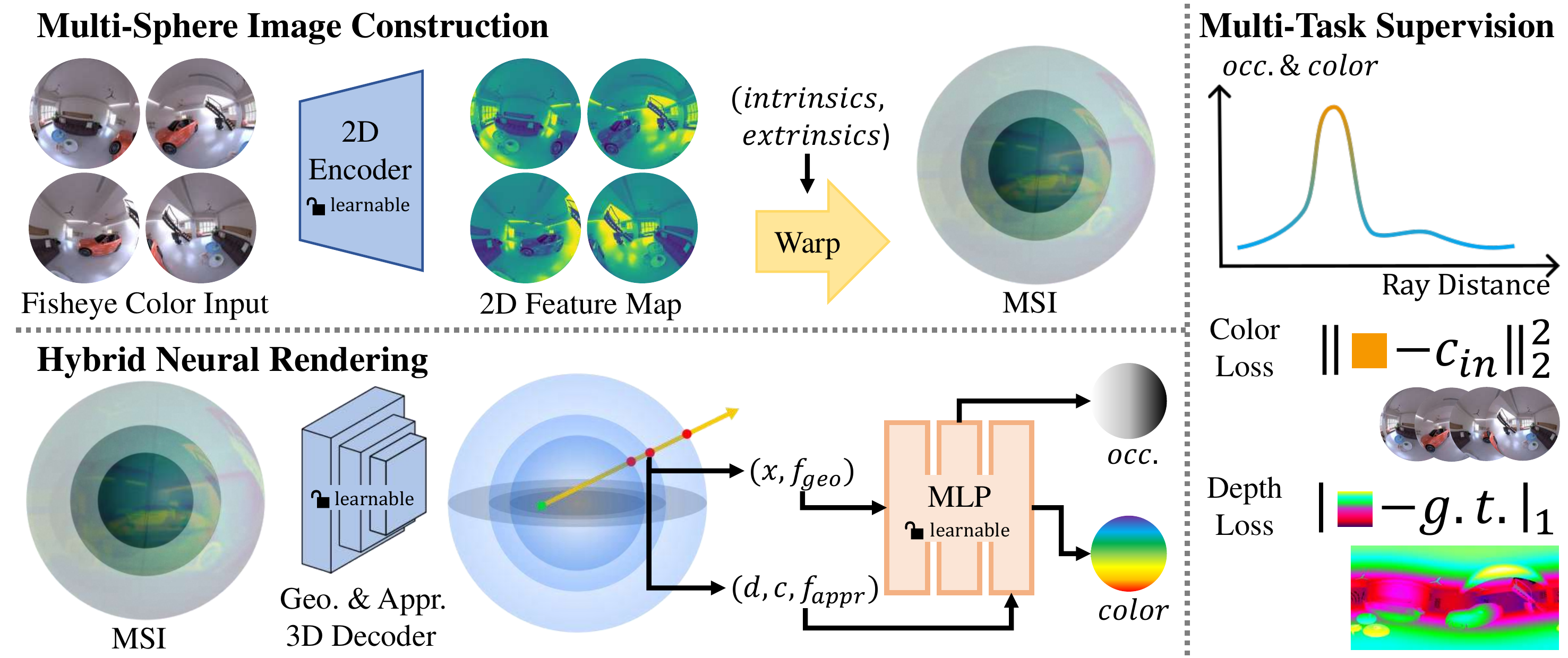}
    \caption{
        Structure of our method.
        Our method can be divided into three parts.
        First, by 2D feature extraction and warping, a multi-sphere image (MSI) representation can be built.
        Then, through geometric and appearance 3D decoders, explicit features ($\mathbf{f}_{geo}$, $\mathbf{f}_{appr}$) containing prior information can be obtained.
        They are then fed into NeRF implicit MLP along with point position $\mathbf{x}$, ray direction $\mathbf{d}$, and projected color $\mathbf{c}$.
        The output occupancy and color are used for fisheye color image and depth image rendering, forming the final supervision loss. 
    }
    \label{fig:structure}
\end{figure*}

\subsection{Generalizable NeRF}

NeRF and its derived algorithms have shown great performance in 3D scene representation.
However, most of them require scene-specific training.
Whenever faced with a new scenario, these methods need seconds or even hours of training before they can generate high-quality results \cite{mildenhall2021nerf, muller2022instant, chen2022tensorf}.
The nature of training from scratch also makes them dependent on dense source views to ensure that novel views can be correctly interpolated.

Leveraging the development of deep learning and stereo matching, NeRF with generalizability gradually emerged \cite{yu2021pixelnerf, wang2021ibrnet, chen2021mvsnerf, wang2022generalizable, fu2024omni, cong2023enhancing, lee2023dense, liang2024retr}.
These generalizable NeRFs eliminate scene-specific training and can quickly synthesize novel views after training on large-scale datasets, even with only sparse view inputs.
PixelNeRF \cite{yu2021pixelnerf} first uses a CNN encoder to extract image features from source views.
It then builds an implicit function that takes the projected 2D features as additional input, following the idea of image-based rendering.
IBRNet \cite{wang2021ibrnet} proposes a similar way to achieve generalizability.
It uses a ray transformer to add interaction along the ray during the render process.
Differently, MVSNeRF \cite{chen2021mvsnerf} incorporates MVS methods into NeRF rendering.
It first constructs a cost volume from the source views using plane sweeping and 3D convolution.
Interpolated 3D features and the projected colors are then added to the NeRF input, transferring prior knowledge into the implicit field.

Although the above methods achieve fast generalizable NeRF reconstruction using only a small number of views, they are all targeted at the object level or forward-facing scene situations.
Obtaining omnidirectional scene representation using a single shot remains a challenging problem.

\section{Method}

In this work, we present MSI-NeRF, a deep learning method that inputs four fisheye images and outputs a panoramic scene representation.
It adopts the representation form of NeRF and uses priorly generated multi-sphere images as a geometric aid.
Such representation enables 6DoF view synthesis while retaining depth information.
During optimization, our method only requires depth ground truth.
It can introduce color information into the network through input data to achieve multi-task supervision.
Our MSI-NeRF consists of three blocks: multi-sphere image construction (Sec. \ref{sec:MSI}), hybrid neural rendering (Sec. \ref{sec:HNR}), and multi-task supervision (Sec. \ref{sec:MTSSS}).
Its overall structure is shown in Fig. \ref{fig:structure}.

\subsection{Multi-Sphere Image Construction}
\label{sec:MSI}

Our method takes four wide FoV fisheye images as input.
They are captured from a multi-fisheye camera rig with four cameras surrounding the $360^\circ$ of the scene.
First, a 2D CNN network with sharing weights extracts feature maps from the captured images.
Then, a multi-sphere image is built to aggregate the multi-view images.
It consists of a series of predefined depth layers.
Each layer is parameterized by equirectangular projection with the same resolution.
In order to handle scenes with infinite depth while considering the reconstruction precision of close-range scenes, we uniformly divide these layers by their inverse depth.
Similar to the plane sweeping methods done in the forward-facing scenes, we project pixels on the MSI back to every fisheye camera using their intrinsics and extrinsics.
The value of each pixel will be filled with the concatenation of the feature vectors from the four feature maps.
The projection is shown in the following equations:
\begin{equation}
    \overline{\mathbf{p}}(\theta, \phi)=(\cos (\phi) \cos (\theta), \sin (\phi), \cos (\phi) \sin (\theta))^{\top},
\end{equation}
\begin{equation}
    \mathbf{V}_{n, k}(\theta, \phi)=F_k\left(\Pi_k\left(\overline{\mathbf{p}}(\theta, \phi) / d^{-1}_n\right)\right),
\end{equation}
where $\overline{\mathbf{p}}$ represents the unit ray for the spherical coordinate, $\theta, \phi$ are the corresponding azimuth and elevation, $\Pi_k$ is the projection function from rig coordinate to the $k^{th}$ camera, and $d^{-1}_n$ is the inverse depth of the $n^{th}$ layer.
$\mathbf{V}$ alias the generated spherical volume representation, and $F$ is the image feature map.

Finally, the constructed spherical layers $\mathbf{V}$ will be processed by 3D CNN networks to match and fuse the features from different viewing angles.
Unlike OmniMVS \cite{won2019omnimvs}, which directly regress the cost volume in the depth dimension to obtain a depth map, and MatryODShka \cite{attal2020matryodshka}, which directly blends colors between these layers, we aim to build a continuous 3D representation that contains both geometry and appearance information.
Specifically, we use two separate networks to further process the obtained MSI volume $\mathbf{V}$.
The first network works as a comparator, generating a cost volume to store geometric information.
The second network additionally outputs an appearance volume, retaining the projected color information.
Their functionality can be demonstrated as the following process:
\begin{equation}
    \mathbf{V}_{geo} = h_{geo}(\mathbf{V}), \quad \mathbf{V}_{app} = h_{app}(\mathbf{V}),
\end{equation}
where $h_{geo}$ and $h_{app}$ represents the two decoders, and $\mathbf{V}_{geo}$ and $\mathbf{V}_{app}$ are the generated feature volumes.
They encode the spatial occupancy and color information respectively, and can be used as explicit aids for radiance field construction and neural rendering.
By distinguishing between appearance and geometry volume representations, we can make our decoder network functions more concrete, each performing their own duties.
It also allows interaction and clarity of network learning.

\subsection{Hybrid Neural Rendering}
\label{sec:HNR}

In order to achieve free viewpoint roaming, continuous representation is needed to infer color and geometric attributes at any location.
We utilize the cutting-edge NeRF \cite{mildenhall2021nerf} for viewpoint interpolation and rendering.
However, vanilla NeRF lacks the generalization ability and can only preserve knowledge of a single scene.
In this work, we incorporate generalizable knowledge from front-end deep neural networks trained on large-scale datasets with NeRF implicit function and form a hybrid neural rendering system.

We construct our implicit function using an MLP.
It can be represented as:
\begin{equation}
o,\mathbf{c}=f(\mathbf{x}, \mathbf{d}, \mathbf{f}_{geo}, \mathbf{f}_{appr}, \mathbf{c}_{proj}, \Theta),
\end{equation}
where $o$ and $\mathbf{c}$ are the outputted occupancy probability and RGB color, $\Theta$ is the network parameter that implicitly saves the scene information, and $\mathbf{x}$ and $\mathbf{d}$ represent the spatial coordinate and viewing direction respectively.
The interpolated geometry and appearance features $\mathbf{f}_{geo}$ and $\mathbf{f}_{appr}$
carrying prior learned information from the MSI volume $\mathbf{V}_{geo}$ and $\mathbf{V}_{app}$ are also fed into the function.
We additionally consider the projected pixel colors $\mathbf{c}_{proj}$ from the original fisheye images as input, providing a hint for color learning.
It can also be viewed as a way of implicit color blending.
The above variables are obtained by the following equation:
\begin{equation}
\begin{aligned}
    \mathbf{f}_{geo}, \mathbf{f}_{appr} &= S(\mathbf{x}, \mathbf{V}_{geo}, \mathbf{V}_{app}), \\
    \mathbf{c}_{proj,k} &= I_k\left(\Pi_k\left(\mathbf{x}\right)\right),
\end{aligned}
\end{equation}
where $S$ represents the interpolation operation, and $I_k$ is the $k^{th}$ color image.
In particular, we first apply positional encoding on the position and direction vectors.
Then, we use an MLP taking embedded $\mathbf{x}$, $\mathbf{d}$, and $\mathbf{f}_{geo}$ as input,
generating the occupancy and a feature vector.
Finally, the feature vector is then processed by another MLP along with $\mathbf{f}_{appr}$ and $\mathbf{c}_{proj}$, generating the final color output.

Since we use inverse depth as the MSI layer arrangement criterion, for distant layers, the distance between two adjacent layers is also greater.
In this way, using 3D interpolation will cause wrongly mixed prior features and affect convergence during training.
As a result, we use the intersection of rays and spherical depth layers as our sample points.
This way, only 2D interpolation on the 3D volume is needed, avoiding blending between depth layers with large gaps while maintaining balanced sample points.
The intersection is calculated by solving a quadratic equation:
\begin{equation}
    a = \mathbf{r}_d^T \mathbf{r}_d, \quad
    b = 2 \mathbf{r}_d^T \mathbf{r}_o, \quad
    c_n = \mathbf{r}_o^T \mathbf{r}_o - 1/(d^{-1}_n)^2,
\end{equation}
where $a$, $b$, and $c$ are the corresponding coefficients, and $\mathbf{r}_o$ and $\mathbf{r}_d$ are the origin and unit direction of the ray $\mathbf{r}$.
The solution that conforms to the perspective relationship can be obtained by the root-finding formula:
\begin{equation}
    z_n = \frac{-b + \sqrt{b^2-4 a c_n}}{2 a} \quad \text{if} \quad \left\|\mathbf{r}_o\right\|_2 < 1/(d^{-1}_n),
\end{equation}
where $z_n$ is the depth of the intersecting point with the $n^{th}$ sphere along the ray.
After points sampling, we follow the pipeline of UNISURF \cite{oechsle2021unisurf} to conduct volume rendering:
\begin{equation}
    \hat{\mathbf{C}}(\mathbf{r})=\sum_{i=1}^{N_s} w\left(\mathbf{p}_i\right)\mathbf{c}\left(\mathbf{p}_i, \mathbf{r}\right), \quad
    \hat{D}(\mathbf{r})=\sum_{i=1}^{N_s}
    w\left(\mathbf{p}_i\right)d\left(\mathbf{p}_i\right),
    \label{eq:render}
\end{equation}
where $N_s$ is the number of intersecting spheres, $\mathbf{p}_i$ denotes the $i^{th}$ sampled points on ray $\mathbf{r}$, and $d(\mathbf{p}_i)$ denotes the depth value at position $\mathbf{p}_i$.
The weight $w\left(\mathbf{p}_i\right)$ is the product of the occupancy value $o(\mathbf{p}_i)$ and the occlusion term $T\left(\mathbf{p}_i\right)$:
\begin{equation}
    w\left(\mathbf{p}_i\right)=o\left(\mathbf{p}_i\right)T\left(\mathbf{p}_i\right), \quad
    T\left(\mathbf{p}_i\right)=\prod_{j=1}^{i-1}\left(1-o\left(\mathbf{p}_j\right)\right).
    \label{eq:weight}
\end{equation}
Through the volume rendering, we can synthesize color images $\hat{\mathbf{C}}(\mathbf{r})$ and depth maps $\hat{D}(\mathbf{r})$ at any viewpoint with any camera intrinsics.

\subsection{Multi-Task Supervision}
\label{sec:MTSSS}

In order to let our network generalize across different scenes, training on large-scale panoramic image datasets is needed.
However, large-scale datasets containing both source fisheye images and multi-view target images are difficult to obtain.
Such difficulties will lead to trouble in learning scene appearance and novel view synthesis.
In this work, we utilize the widely used
fisheye omnidirectional depth estimation datasets proposed by OmniMVS \cite{won2019omnimvs} for both geometry and appearance understanding of panoramic scenes.
We revisit the training mechanism of generalizable NeRFs and find that the purpose of using multi-view target images for network training is: \textbf{(a)} to predict scene occupy information and \textbf{(b)} to predict the appearance information of the occluded part of the scene.
These two points allow generalizable NeRFs to quickly synthesize novel views from a small number of target views, using the learned prior knowledge.
Now that we already have the depth supervision to learn occupancy, we only need to find a way to inject occluded color supervision into training.
We come up with the idea of directly using the input fisheye images as a multi-view supervision to provide texture information for the occluded parts.

To this end, we propose our multi-task supervision strategy.
For the original depth ground truth aligned with the rig coordinate, we use the rendered depth to compare it as depth supervision.
We further render fisheye color images using the source camera's intrinsic and extrinsic and compare them with the captured images, forming an additional color supervision.
Our loss function is represented as:
\begin{equation}
\begin{aligned}
    \mathcal{L}_{depth}&=\frac{1}{|\mathcal{R}_p|}\sum_{\mathbf{r} \in \mathcal{R}_p}\left\|\hat{D}(\mathbf{r})-D(\mathbf{r})\right\|_1, \\
    \mathcal{L}_{color}&=\frac{1}{|\mathcal{R}_f|}\sum_{\mathbf{r} \in \mathcal{R}_f}\left\|\hat{\mathbf{C}}(\mathbf{r})-\mathbf{C}(\mathbf{r})\right\|_2^2,
\end{aligned}
\end{equation}
\begin{equation}
    \mathcal{L}=\mathcal{L}_{color}+\lambda_d \mathcal{L}_{depth},
\end{equation}
where $\mathcal{R}_p$ is the set of rays that makes up the central panoramic depth image, and $\mathcal{R}_f$ is the set of rays that constitutes the four fisheye images.

The depth supervision can provide a useful geometry guidance to the multi-sphere images, which also shares knowledge to appearance network and benefit color image generation.
If only source fisheye images are used for training, turning the method into fully self-supervised, the network will suffer from insufficient overlapping and bad scene structure prediction.
Adding depth supervision greatly constrained such problems.

\section{Experiments}

\begin{figure*}[t]
    \centering
    \includegraphics[width=0.9\textwidth]{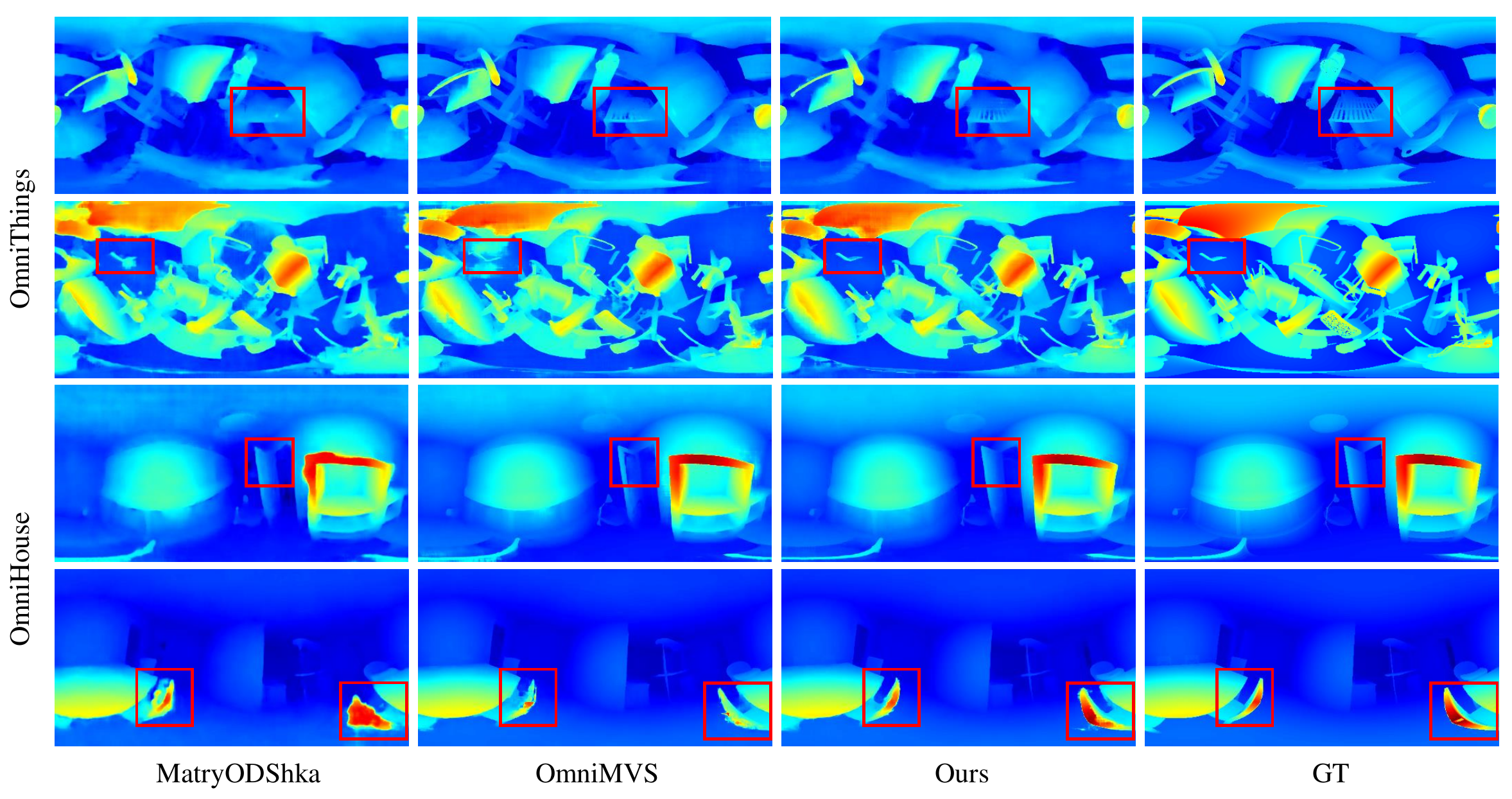}
    \caption{
        Qualitative results of our depth estimation experiment.
        We compare the generated omnidirectional depth map with the ground truth depth from the dataset.
        Our method can generate fine-grained depth estimations while getting rid of over-fitting due to rich texture.
    }
    \label{fig:depth}
\end{figure*}

\begin{table*}[h]
\tiny
\centering
\resizebox{\linewidth}{!}{
\renewcommand\arraystretch{1.2}
\begin{tabular}{ccccccccccc}
\hline
\textbf{Dataset} & \multicolumn{5}{c}{\textbf{OmniHouse}} & \multicolumn{5}{c}{\textbf{OmniThings}}  \\
\hline
\textbf{Method}       & $>$ \textbf{0.1} & $>$ \textbf{0.3} & $>$ \textbf{0.5} & \textbf{MAE} & \textbf{RMSE}  & $>$ \textbf{0.1} & $>$ \textbf{0.3} & $>$ \textbf{0.5} & \textbf{MAE} & \textbf{RMSE}  \\
\hline
\textbf{MatryODShka}  & 2.091          & 0.271            & \underline{0.066}           & 0.017          & 0.032  & 12.989          & 4.046            & 1.662           & 0.053        & 0.032 \\
\textbf{OmniMVS}      & \textbf{1.253}          & \textbf{0.117}            & 0.071           & \textbf{0.011}          & \textbf{0.023}  & \underline{10.592}          & \underline{3.265}            & \textbf{1.305}           & \underline{0.046}          & \underline{0.105}   \\
\textbf{Ours}         & \underline{1.255}          & \underline{0.180}            & \textbf{0.033}           & \textbf{0.011}          & \underline{0.024}  & \textbf{10.149}          & \textbf{3.212}            & \underline{1.328}           & \textbf{0.044}          & \textbf{0.103}  \\
\hline
\end{tabular}}
\caption{
    Quantitative evaluation result of depth estimation.
  }
\label{tab:depth}
\end{table*}

\subsection{Implementation Details}

We implement our method using the Pytorch framework.
Training and validation are all carried out on a GPU server with four RTX 2080Ti.
During optimization, we use ADAM \cite{kingma2014adam} optimizer with a learning rate of $3 \times 10^{-4}$.
The loss weight of depth is set to $\lambda_d = 5$.
We first train the base model on OmniThings dataset for $30$ epochs and then fine-tune it on OmniHouse dataset for $10$ \cite{won2019omnimvs}.
We set the inverse depth layer number to $64$ and the range of layer value to $\left[0,2\right] m^{-1}$.
Due to limited GPU memory, after inference of geometry and appearance volumes, we randomly sample $8192$ rays on fisheyes and $8192$ rays on the omnidirectional depth map, expressed as $|\mathcal{R}_p|=|\mathcal{R}_f|=8192$.
Then, volume rendering is done on these rays for further color and depth supervision.

\subsection{Datasets}

We use the fisheye panorama datasets proposed in OmniMVS \cite{won2019omnimvs} for network training.
Specifically, we use OmniThings for base model training and OmniHouse for indoor scene fine-tuning.
In addition to the four surrounding fisheye cameras, each sample contains a depth ground-truth image in the center.
The last $5\%$ of the dataset is extracted for validation.

\begin{figure}[h]
    \centering
    \includegraphics[width=0.45\textwidth]{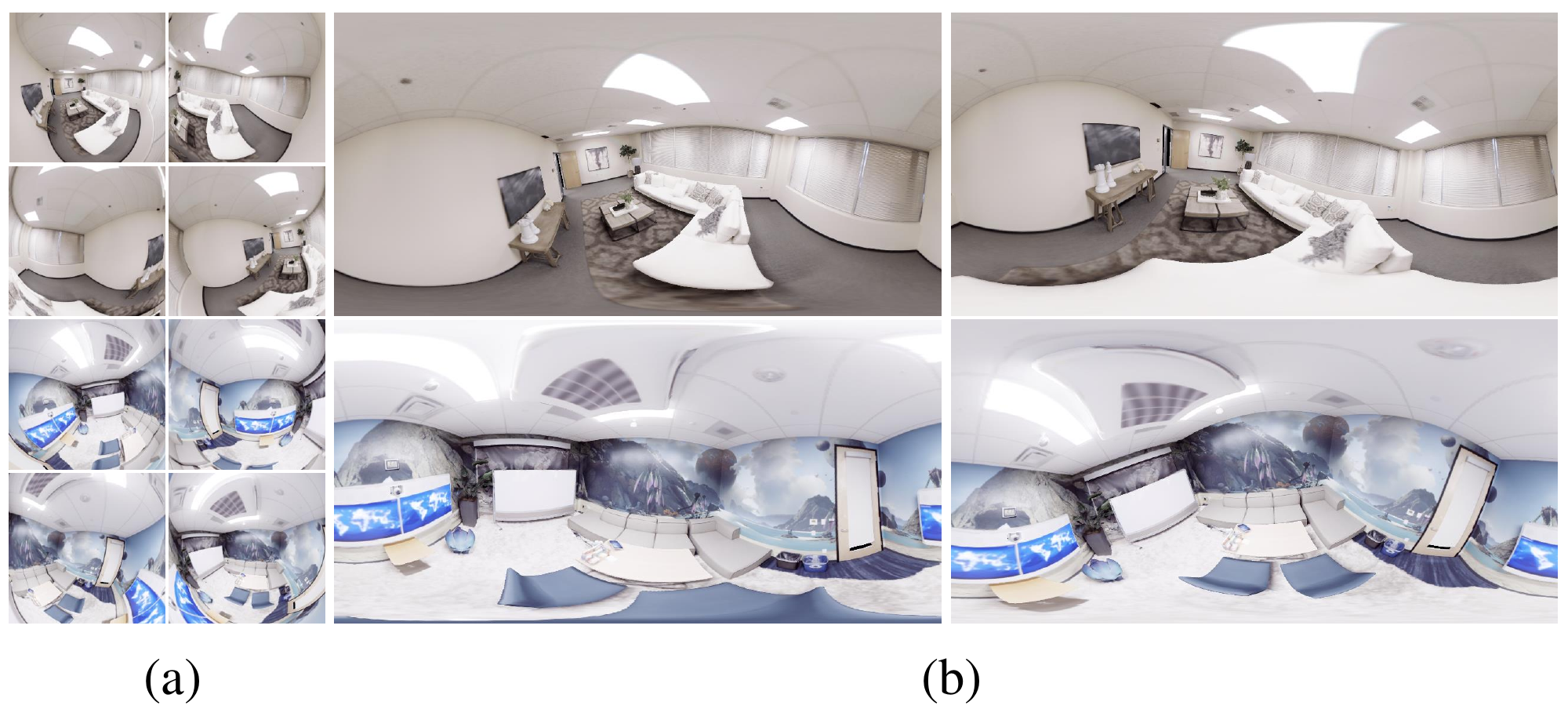}
    \caption{
        Visualization of our Replica360 dataset.
        The fisheye source views \textbf{(a)}, and the multi-view omnidirectional target images \textbf{(b)} are captured from the Replica simulator.
    }
\label{fig:dataset}
\end{figure}

We also capture fisheye and multi-view images in the Replica \cite{straub2019replica} simulator to verify the novel view synthesis performance.
The Replica360 dataset we constructed contains $33$ indoor scenes, each with $4$ fisheye source views and $5$ randomly sampled omnidirectional target views.
The visualization of our Replica360 dataset is in Fig. \ref{fig:dataset}.

\begin{figure*}[t]
    \centering
    \includegraphics[width=0.9\textwidth]{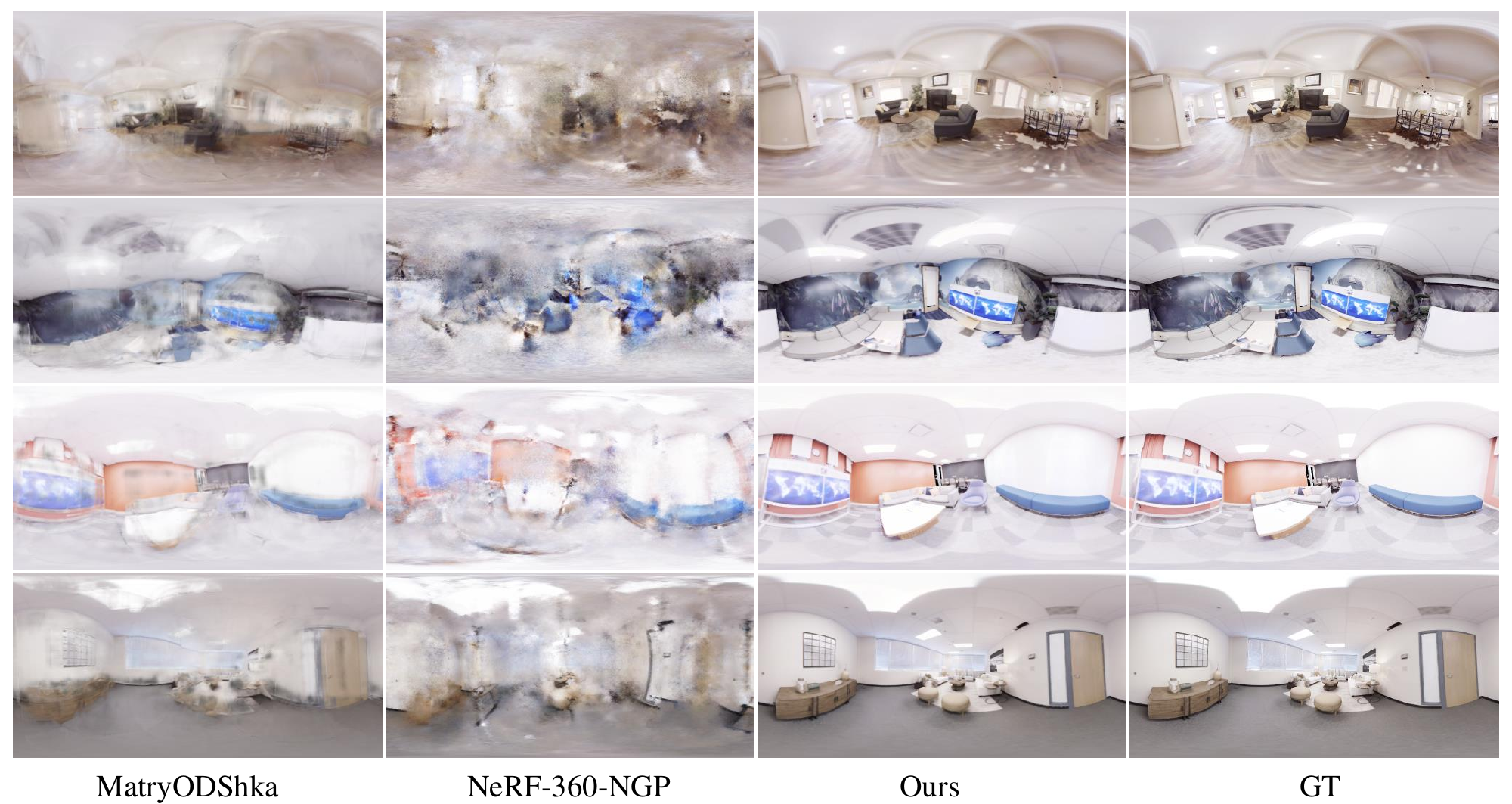}
    \caption{        
        Qualitative results of our novel view synthesis experiment.
        We generate novel panoramic images in the dataset's target view location and compare them with the ground truth.
        Our method can generate high-quality and consistent rendering results, avoiding blurring and ghosting artifacts.
    }
    \label{fig:nvs}
\end{figure*}

\subsection{Depth Evaluation}

We evaluate the omnidirectional depth estimation performance on a validation set extracted from the OmniHouse and OmniThings.
Original and fine-tuned models are used for different datasets, respectively.
We reimplement a compressed version of OmniMVS \cite{won2019omnimvs} using the same backbone as ours, keeping the network capacity and hyperparameters consistent with ours.
The qualitative and quantitative evaluation results are shown in Fig. \ref{fig:depth} and Tab. \ref{tab:depth}.
We use inverse depth value as our comparison metric, and the unit is set to $m^{-1}$.
The qualifier "$>k$" represents the pixel ratio whose MAE is larger than threshold $k$.

It can be seen that the depth map generated by MatryODShka \cite{attal2020matryodshka} has blurring effect and poor accuracy.
Although OmniMVS \cite{won2019omnimvs} can generate fine object edges, it focuses too much on texture, resulting in incorrect depth estimation on surfaces with rich textures.
Our method is able to generate fine edges while maintaining overall accuracy.

\subsection{Novel View Synthesis Evaluation}

To test the novel view synthesis ability, we conduct comparative experiments against MatryODShka \cite{attal2020matryodshka} and NeRF-360-NGP \cite{mildenhall2021nerf, barron2022mip, muller2022instant}.
Since MatryODShka uses binocular omnidirectional stereo (ODS) images as input which is also originated from fisheye cameras, we re-implement it using our fisheye input setting to align the input format.
In particular, we use sphere sweep method directly on the input fisheye images to generate four sphere sweep volumes (SSV) and pass them to the network to get the blending weights.
We also add a depth renderer so that it can be trained using depth ground-truth only.
As for NeRF-360-NGP, we expand the NeRF \cite{mildenhall2021nerf} method to $360^\circ$ using spatial parameterization from Mip-NeRF 360 \cite{barron2022mip} and achieve efficient training through Instant-NGP \cite{muller2022instant}.

We show our qualitative and quantitative results in Fig. \ref{fig:nvs} and Tab. \ref{tab:nvs}.
It can be seen that the color blending design in MatryODShka generates ghosting effect when trained with only depth ground truth.
NeRF-360-NGP fails to generate reasonable novel views with limited input images and shared viewing angles.
Our method exceeds other generalizable and scene-specific methods in terms of image quality.
It achieves high accuracy on novel view synthesis and shows excellent generalization ability on cross-dataset validation.

\begin{table}[h]
\resizebox{\linewidth}{!}{
\renewcommand\arraystretch{1.2}
\begin{tabular}{cccccc}
\hline
\textbf{Method}   &  \textbf{Generalizable}  & \textbf{PSNR} & \textbf{SSIM} & \textbf{LPIPS} & \textbf{Time}  \\
\hline
\textbf{MatryODShka} & \CheckmarkBold   & \underline{29.665}          & \underline{0.584}            & \underline{0.419}           &
\textbf{0.128}s                 \\
\textbf{NeRF-360-NGP}  & \color{red}{\XSolidBrush} & 28.594          & 0.451            & 0.541      & $\sim$10s     \\
\textbf{Ours} & \CheckmarkBold  & \textbf{37.302}          & \textbf{0.957}            & \textbf{0.077}  & \underline{0.383}s          \\
\hline
\end{tabular}}
\centering
  \caption{
    Quantitative evaluation result of novel view synthesis.
  }
\label{tab:nvs}
\end{table}

\begin{figure*}[h]
    \centering
    \includegraphics[width=0.9\textwidth]{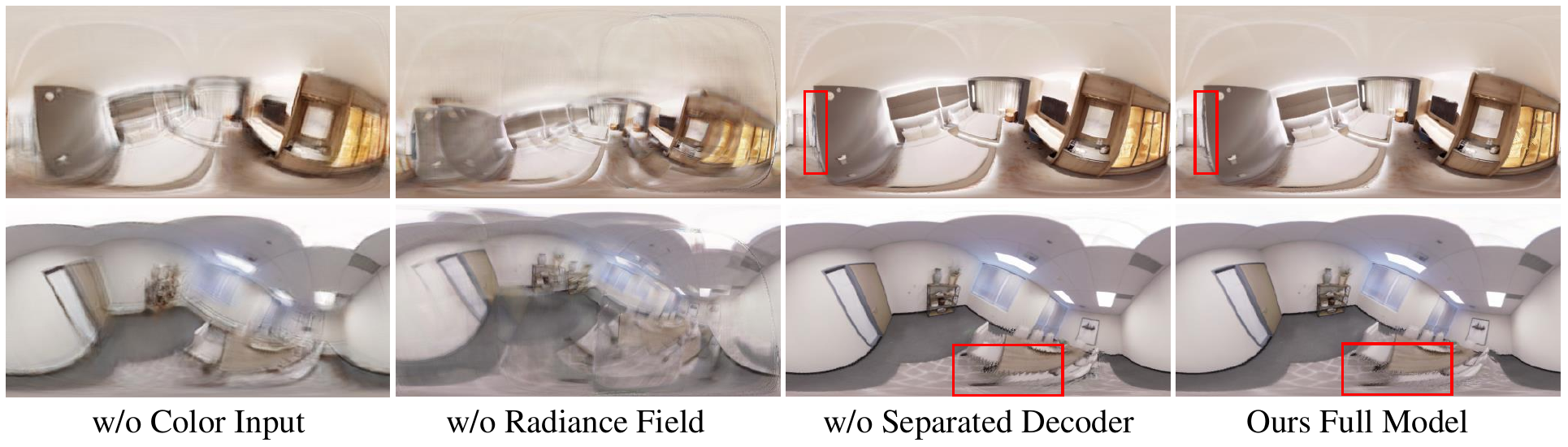}
    \caption{
        Qualitative results of our ablation studies.
        We generate novel view images of our method without key designs and compare them with our full model.
        Ours full model generates the novel view images with the best overall quality and finest details.
    }
    \label{fig:ablation}
\end{figure*}

\subsection{Ablation Studies}

Our generalizable neural rendering method does not follow traditional novel view synthesis process.
It is mainly reflected in three key designs: additional color projection, radiance field rendering, and separated geometry and appearance decoder.
We conduct ablation studies to demonstrate the effect of these designs by removing them individually.
The results are shown in Fig. \ref{fig:ablation} and Tab. \ref{tab:ablation}.

The method without additional color input struggles at color learning, producing blurry images.
The method without radiance field rendering directly outputs color and occupancy spheres, causing ghosting effects in the synthesized views.
The method without separated decoders lacks the ability to estimate fine-grained depth maps.
Ours full model uses color projection input to gain additional color information and implicitly mix them to achieve more reasonable color output.
The usage of radiance field rendering creates a more consistent image generation and results in finer color and depth generation.
Moreover, the design of separated geometry and appearance decoder makes the role of the network clearer and avoids data confusion.

\begin{figure}[htpb]
    \centering
    \includegraphics[width=0.5\textwidth]{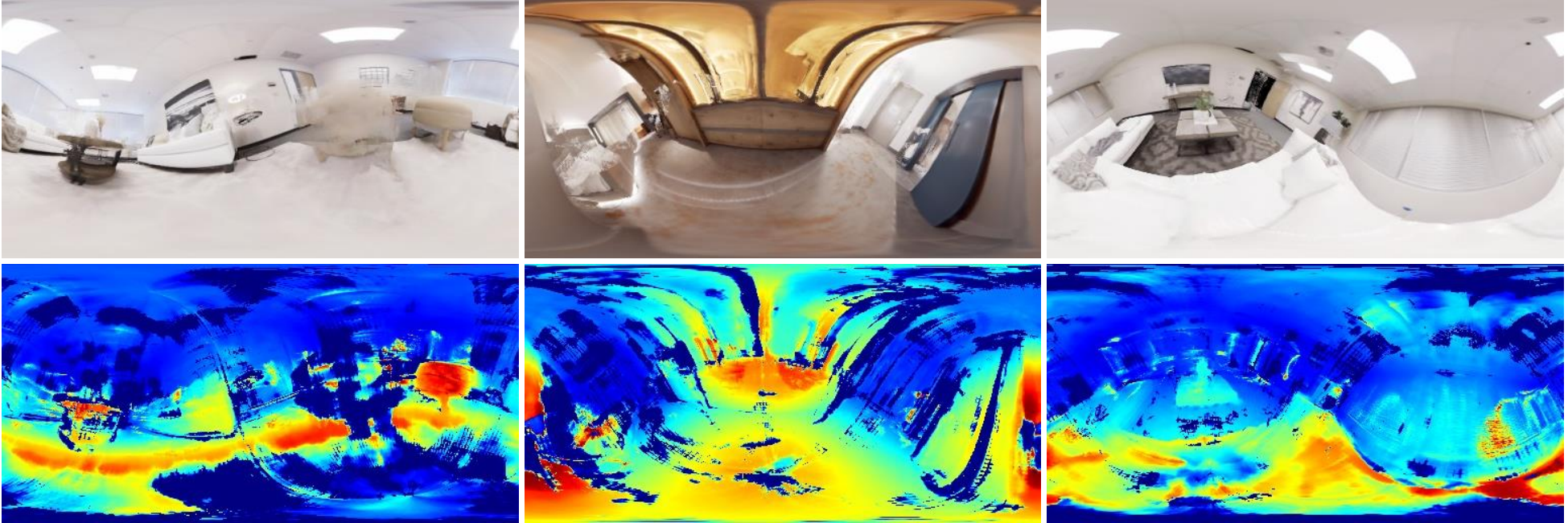}
    \caption{
        Visualization of our method with only color supervision 
    }
\label{fig:nodepth}
\end{figure}

We also test our method with only color image for supervision (self-supervised), and the results are shown in Fig. \ref{fig:nodepth}.
Although the structure of explicit MSI volumes along with implicit NeRF rendering works as geometry proxy and reasonable rendering results can be generated, they are of severe artifacts which are far from being able to be applied in upper-level applications.

\begin{figure}[htpb]
    \centering
    \includegraphics[width=0.5\textwidth]{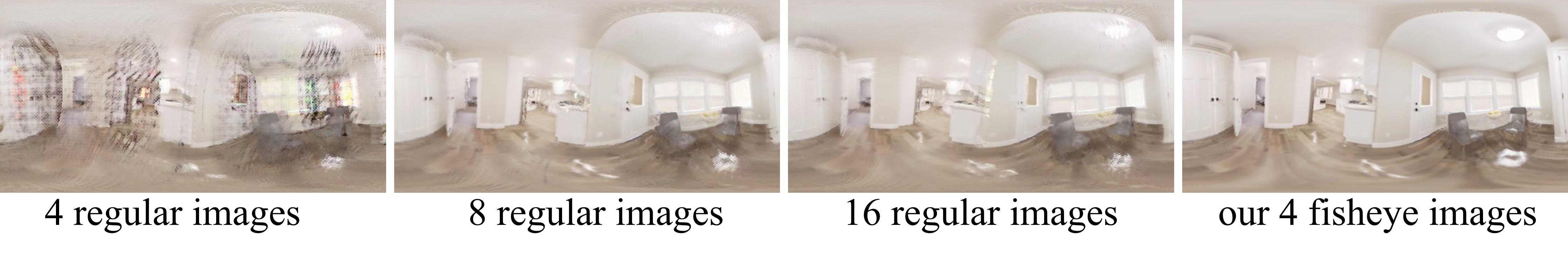}
    \caption{
        Visualization of our fisheye image input with regular image input.
    }
\label{fig:nodepth}
\end{figure}

To highlight the advantage of our fisheye input format, we compare our method against the method that uses regular image input.
We reimplement our method using regular image input by warping the input to the MSI structure.
The result shows that more than 16 regular images are needed to achieve similar reconstruction quality.

\begin{table}[h]
\tiny
\centering
\resizebox{\linewidth}{!}{
\renewcommand\arraystretch{1.2}
\begin{tabular}{ccccccccccc}
\hline
\textbf{Method} & \textbf{PSNR} & \textbf{SSIM} & \textbf{LPIPS}  \\
\hline
\textbf{w/o Color Input} & 32.080          & 0.783            & 0.287           \\
\textbf{w/o Radiance Field} & 29.620          & 0.553            & 0.432           \\
\textbf{w/o Separated Decoder} & \underline{37.200}          & \underline{0.953}            & \underline{0.088}           \\
\textbf{Ours Full Model} &\textbf{37.302}          & \textbf{0.957}            & \textbf{0.077}           \\
\hline
\end{tabular}}
\caption{
    Quantitative evaluation result of ablation studies.
  }
\label{tab:ablation}
\end{table}

\subsection{Real-world Experiment}

\begin{figure}[htpb]
    \centering
    \includegraphics[width=0.5\textwidth]{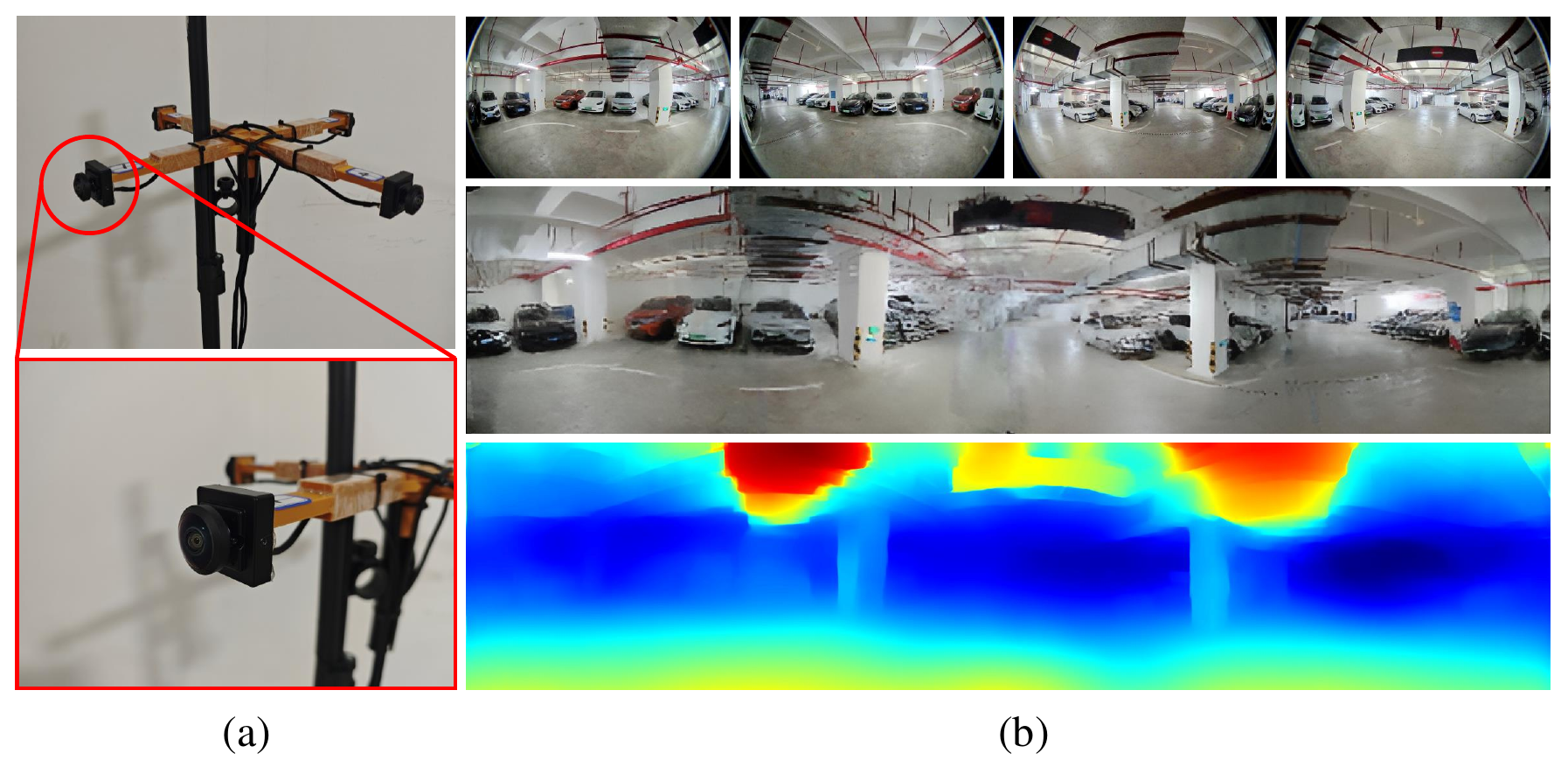}
    \caption{
        \textbf{(a)}: Our self-made multi-fisheye camera rig system for real-world data capture.
        It comprises $4$ fisheye cameras with $220^\circ$ FoV arranged on a circle with a diameter of $0.4m$.
        \textbf{(b)}: The captured source views, the generated omnidirectional depth map, and the synthesized image in the real-world experiment.
    }
\label{fig:real_world}
\end{figure}

To test our approach's generalizability and real-world performance, we constructed a self-collected dataset using a multi-fisheye camera rig with similar settings to the synthetic datasets.
The data capture system, fisheye image samples, and the generated results can be seen in Fig. \ref{fig:real_world}.
From the generated depth and novel view images, we can see that our method can generalize from sim to real.
The real-world experiment also demonstrates the practical application value for virtual reality video shooting and 3D reconstruction in real-world scenes.
However, limitations exist because it requires accurate camera intrinsic and extrinsic parameters and calibration errors may cause blurring and artifacts in color and depth images in real-world scenes.

\section{Conclusion and Limitation}

This paper explores the problem of generating a panoramic scene representation using only four fisheye images as input.
Our presented MSI-NeRF combines the advantage of MSI representation with volume rendering to construct a generalizable NeRF.
From such representation, accurate omni-depth estimation and high-quality novel view synthesis can be conducted efficiently, with only a single network inference.
Our method outperforms baseline methods in both image quality and feasibility, which is significant for virtual reality and robotics applications.
Since our method uses the idea of MVS and does not have generation ability, the reconstruction of the occluded part is unpleasant.
In the future research, we hope to leverage the power of generative models to inpaint the unseen part.

\clearpage
{\small
\bibliographystyle{ieee_fullname}
\bibliography{egbib}

\begin{thebibliography}{10}\itemsep=-1pt

\bibitem{attal2020matryodshka}
Benjamin Attal, Selena Ling, Aaron Gokaslan, Christian Richardt, and James Tompkin.
\newblock Matryodshka: Real-time 6dof video view synthesis using multi-sphere images.
\newblock In {\em European Conference on Computer Vision}, pages 441--459. Springer, 2020.

\bibitem{barron2021mip}
Jonathan~T Barron, Ben Mildenhall, Matthew Tancik, Peter Hedman, Ricardo Martin-Brualla, and Pratul~P Srinivasan.
\newblock Mip-nerf: A multiscale representation for anti-aliasing neural radiance fields.
\newblock In {\em Proceedings of the IEEE/CVF International Conference on Computer Vision}, pages 5855--5864, 2021.

\bibitem{barron2022mip}
Jonathan~T Barron, Ben Mildenhall, Dor Verbin, Pratul~P Srinivasan, and Peter Hedman.
\newblock Mip-nerf 360: Unbounded anti-aliased neural radiance fields.
\newblock In {\em Proceedings of the IEEE/CVF Conference on Computer Vision and Pattern Recognition}, pages 5470--5479, 2022.

\bibitem{barron2023zip}
Jonathan~T Barron, Ben Mildenhall, Dor Verbin, Pratul~P Srinivasan, and Peter Hedman.
\newblock Zip-nerf: Anti-aliased grid-based neural radiance fields.
\newblock {\em arXiv preprint arXiv:2304.06706}, 2023.

\bibitem{brown2007automatic}
Matthew Brown and David~G Lowe.
\newblock Automatic panoramic image stitching using invariant features.
\newblock {\em International journal of computer vision}, 74:59--73, 2007.

\bibitem{broxton2019low}
Michael Broxton, Jay Busch, Jason Dourgarian, Matthew DuVall, Daniel Erickson, Dan Evangelakos, John Flynn, Ryan Overbeck, Matt Whalen, and Paul Debevec.
\newblock A low cost multi-camera array for panoramic light field video capture.
\newblock In {\em SIGGRAPH Asia 2019 Posters}, pages 1--2. 2019.

\bibitem{chang2018pyramid}
Jia-Ren Chang and Yong-Sheng Chen.
\newblock Pyramid stereo matching network.
\newblock In {\em Proceedings of the IEEE conference on computer vision and pattern recognition}, pages 5410--5418, 2018.

\bibitem{chen2022tensorf}
Anpei Chen, Zexiang Xu, Andreas Geiger, Jingyi Yu, and Hao Su.
\newblock Tensorf: Tensorial radiance fields.
\newblock In {\em European Conference on Computer Vision}, pages 333--350. Springer, 2022.

\bibitem{chen2021mvsnerf}
Anpei Chen, Zexiang Xu, Fuqiang Zhao, Xiaoshuai Zhang, Fanbo Xiang, Jingyi Yu, and Hao Su.
\newblock Mvsnerf: Fast generalizable radiance field reconstruction from multi-view stereo.
\newblock In {\em Proceedings of the IEEE/CVF International Conference on Computer Vision}, pages 14124--14133, 2021.

\bibitem{chen2023unsupervised}
Zisong Chen, Chunyu Lin, Lang Nie, Kang Liao, and Yao Zhao.
\newblock Unsupervised omnimvs: Efficient omnidirectional depth inference via establishing pseudo-stereo supervision.
\newblock In {\em 2023 IEEE/RSJ International Conference on Intelligent Robots and Systems (IROS)}, pages 10873--10879. IEEE, 2023.

\bibitem{choi2023balanced}
Changwoon Choi, Sang~Min Kim, and Young~Min Kim.
\newblock Balanced spherical grid for egocentric view synthesis.
\newblock In {\em Proceedings of the IEEE/CVF Conference on Computer Vision and Pattern Recognition}, pages 16590--16599, 2023.

\bibitem{cong2023enhancing}
Wenyan Cong, Hanxue Liang, Peihao Wang, Zhiwen Fan, Tianlong Chen, Mukund Varma, Yi Wang, and Zhangyang Wang.
\newblock Enhancing nerf akin to enhancing llms: Generalizable nerf transformer with mixture-of-view-experts.
\newblock In {\em Proceedings of the IEEE/CVF International Conference on Computer Vision}, pages 3193--3204, 2023.

\bibitem{fu2024omni}
Yonggan Fu, Huaizhi Qu, Zhifan Ye, Chaojian Li, Kevin Zhao, and Yingyan Lin.
\newblock Omni-recon: Towards general-purpose neural radiance fields for versatile 3d applications.
\newblock {\em arXiv preprint arXiv:2403.11131}, 2024.

\bibitem{gao2017dual}
Wenliang Gao and Shaojie Shen.
\newblock Dual-fisheye omnidirectional stereo.
\newblock In {\em 2017 IEEE/RSJ International Conference on Intelligent Robots and Systems (IROS)}, pages 6715--6722. IEEE, 2017.

\bibitem{gao2020autonomous}
Wenliang Gao, Kaixuan Wang, Wenchao Ding, Fei Gao, Tong Qin, and Shaojie Shen.
\newblock Autonomous aerial robot using dual-fisheye cameras.
\newblock {\em Journal of Field Robotics}, 37(4):497--514, 2020.

\bibitem{gu2023vanishing}
Kai Gu, Thomas Maugey, Sebastian Knorr, and Christine Guillemot.
\newblock Vanishing point aided hash-frequency encoding for neural radiance fields (nerf) from sparse 360° input.
\newblock In {\em 2023 IEEE International Symposium on Mixed and Augmented Reality (ISMAR)}, pages 1142--1151. IEEE, 2023.

\bibitem{hane2014real}
Christian H{\"a}ne, Lionel Heng, Gim~Hee Lee, Alexey Sizov, and Marc Pollefeys.
\newblock Real-time direct dense matching on fisheye images using plane-sweeping stereo.
\newblock In {\em 2014 2nd International Conference on 3D Vision}, volume~1, pages 57--64. IEEE, 2014.

\bibitem{hara2022enhancement}
Takayuki Hara and Tatsuya Harada.
\newblock Enhancement of novel view synthesis using omnidirectional image completion.
\newblock {\em arXiv preprint arXiv:2203.09957}, 2022.

\bibitem{hirschmuller2007stereo}
Heiko Hirschmuller.
\newblock Stereo processing by semiglobal matching and mutual information.
\newblock {\em IEEE Transactions on pattern analysis and machine intelligence}, 30(2):328--341, 2007.

\bibitem{abs-2106-10859}
Ching{-}Yu Hsu, Cheng Sun, and Hwann{-}Tzong Chen.
\newblock Moving in a 360 world: Synthesizing panoramic parallaxes from a single panorama.
\newblock {\em CoRR}, abs/2106.10859, 2021.

\bibitem{kingma2014adam}
Diederik~P Kingma and Jimmy Ba.
\newblock Adam: A method for stochastic optimization.
\newblock {\em arXiv preprint arXiv:1412.6980}, 2014.

\bibitem{kulkarni2023360fusionnerf}
Shreyas Kulkarni, Peng Yin, and Sebastian Scherer.
\newblock 360fusionnerf: Panoramic neural radiance fields with joint guidance.
\newblock In {\em 2023 IEEE/RSJ International Conference on Intelligent Robots and Systems (IROS)}, pages 7202--7209. IEEE, 2023.

\bibitem{lee2023dense}
Dongwoo Lee and Kyoung~Mu Lee.
\newblock Dense depth-guided generalizable nerf.
\newblock {\em IEEE Signal Processing Letters}, 30:75--79, 2023.

\bibitem{li2021mine}
Jiaxin Li, Zijian Feng, Qi She, Henghui Ding, Changhu Wang, and Gim~Hee Lee.
\newblock Mine: Towards continuous depth mpi with nerf for novel view synthesis.
\newblock In {\em Proceedings of the IEEE/CVF International Conference on Computer Vision}, pages 12578--12588, 2021.

\bibitem{lian2018image}
Trisha Lian, Joyce Farrell, and Brian Wandell.
\newblock Image systems simulation for 360 camera rigs.
\newblock {\em Electronic Imaging}, 2018(5):353--1, 2018.

\bibitem{liang2024retr}
Yixun Liang, Hao He, and Yingcong Chen.
\newblock Retr: Modeling rendering via transformer for generalizable neural surface reconstruction.
\newblock {\em Advances in Neural Information Processing Systems}, 36, 2024.

\bibitem{liu2024omninxt}
Peize Liu, Chen Feng, Yang Xu, Yan Ning, Hao Xu, and Shaojie Shen.
\newblock Omninxt: A fully open-source and compact aerial robot with omnidirectional visual perception.
\newblock {\em arXiv preprint arXiv:2403.20085}, 2024.

\bibitem{meuleman2021real}
Andreas Meuleman, Hyeonjoong Jang, Daniel~S Jeon, and Min~H Kim.
\newblock Real-time sphere sweeping stereo from multiview fisheye images.
\newblock In {\em Proceedings of the IEEE/CVF Conference on Computer Vision and Pattern Recognition}, pages 11423--11432, 2021.

\bibitem{mildenhall2021nerf}
Ben Mildenhall, Pratul~P Srinivasan, Matthew Tancik, Jonathan~T Barron, Ravi Ramamoorthi, and Ren Ng.
\newblock Nerf: Representing scenes as neural radiance fields for view synthesis.
\newblock {\em Communications of the ACM}, 65(1):99--106, 2021.

\bibitem{muhlhausen2023immersive}
Moritz M{\"u}hlhausen, Moritz Kappel, Marc Kassubeck, Leslie W{\"o}hler, Steve Grogorick, Susana Castillo, Martin Eisemann, and Marcus Magnor.
\newblock Immersive free-viewpoint panorama rendering from omnidirectional stereo video.
\newblock In {\em Computer Graphics Forum}. Wiley Online Library, 2023.

\bibitem{muller2022instant}
Thomas M{\"u}ller, Alex Evans, Christoph Schied, and Alexander Keller.
\newblock Instant neural graphics primitives with a multiresolution hash encoding.
\newblock {\em ACM Transactions on Graphics (ToG)}, 41(4):1--15, 2022.

\bibitem{oechsle2021unisurf}
Michael Oechsle, Songyou Peng, and Andreas Geiger.
\newblock Unisurf: Unifying neural implicit surfaces and radiance fields for multi-view reconstruction.
\newblock In {\em Proceedings of the IEEE/CVF International Conference on Computer Vision}, pages 5589--5599, 2021.

\bibitem{straub2019replica}
Julian Straub, Thomas Whelan, Lingni Ma, Yufan Chen, Erik Wijmans, Simon Green, Jakob~J Engel, Raul Mur-Artal, Carl Ren, Shobhit Verma, et~al.
\newblock The replica dataset: A digital replica of indoor spaces.
\newblock {\em arXiv preprint arXiv:1906.05797}, 2019.

\bibitem{wang2022generalizable}
Dan Wang, Xinrui Cui, Septimiu Salcudean, and Z~Jane Wang.
\newblock Generalizable neural radiance fields for novel view synthesis with transformer.
\newblock {\em arXiv preprint arXiv:2206.05375}, 2022.

\bibitem{wang2023perf}
Guangcong Wang, Peng Wang, Zhaoxi Chen, Wenping Wang, Chen~Change Loy, and Ziwei Liu.
\newblock Perf: Panoramic neural radiance field from a single panorama.
\newblock {\em arXiv preprint arXiv:2310.16831}, 2023.

\bibitem{wang2023f2}
Peng Wang, Yuan Liu, Zhaoxi Chen, Lingjie Liu, Ziwei Liu, Taku Komura, Christian Theobalt, and Wenping Wang.
\newblock F2-nerf: Fast neural radiance field training with free camera trajectories.
\newblock In {\em Proceedings of the IEEE/CVF Conference on Computer Vision and Pattern Recognition}, pages 4150--4159, 2023.

\bibitem{wang2021ibrnet}
Qianqian Wang, Zhicheng Wang, Kyle Genova, Pratul~P Srinivasan, Howard Zhou, Jonathan~T Barron, Ricardo Martin-Brualla, Noah Snavely, and Thomas Funkhouser.
\newblock Ibrnet: Learning multi-view image-based rendering.
\newblock In {\em Proceedings of the IEEE/CVF Conference on Computer Vision and Pattern Recognition}, pages 4690--4699, 2021.

\bibitem{won2019omnimvs}
Changhee Won, Jongbin Ryu, and Jongwoo Lim.
\newblock Omnimvs: End-to-end learning for omnidirectional stereo matching.
\newblock In {\em Proceedings of the IEEE/CVF International Conference on Computer Vision}, pages 8987--8996, 2019.

\bibitem{won2019sweepnet}
Changhee Won, Jongbin Ryu, and Jongwoo Lim.
\newblock Sweepnet: Wide-baseline omnidirectional depth estimation.
\newblock In {\em 2019 International Conference on Robotics and Automation (ICRA)}, pages 6073--6079. IEEE, 2019.

\bibitem{xie2023omnividar}
Sheng Xie, Daochuan Wang, and Yun-Hui Liu.
\newblock Omnividar: Omnidirectional depth estimation from multi-fisheye images.
\newblock In {\em Proceedings of the IEEE/CVF Conference on Computer Vision and Pattern Recognition}, pages 21529--21538, 2023.

\bibitem{xu2022omni}
Hao Xu, Yichen Zhang, Boyu Zhou, Luqi Wang, Xinjie Yao, Guotao Meng, and Shaojie Shen.
\newblock Omni-swarm: A decentralized omnidirectional visual--inertial--uwb state estimation system for aerial swarms.
\newblock {\em IEEE Transactions on Robotics}, 38(6):3374--3394, 2022.

\bibitem{yao2018mvsnet}
Yao Yao, Zixin Luo, Shiwei Li, Tian Fang, and Long Quan.
\newblock Mvsnet: Depth inference for unstructured multi-view stereo.
\newblock In {\em Proceedings of the European conference on computer vision (ECCV)}, pages 767--783, 2018.

\bibitem{yu2021pixelnerf}
Alex Yu, Vickie Ye, Matthew Tancik, and Angjoo Kanazawa.
\newblock pixelnerf: Neural radiance fields from one or few images.
\newblock In {\em Proceedings of the IEEE/CVF Conference on Computer Vision and Pattern Recognition}, pages 4578--4587, 2021.

\bibitem{zhang2020nerf++}
Kai Zhang, Gernot Riegler, Noah Snavely, and Vladlen Koltun.
\newblock Nerf++: Analyzing and improving neural radiance fields.
\newblock {\em arXiv preprint arXiv:2010.07492}, 2020.

\bibitem{zhou2018stereo}
Tinghui Zhou, Richard Tucker, John Flynn, Graham Fyffe, and Noah Snavely.
\newblock Stereo magnification: Learning view synthesis using multiplane images.
\newblock {\em arXiv preprint arXiv:1805.09817}, 2018.

\bibitem{zuckerberg2016360}
Mark Zuckerberg.
\newblock 360 video: Surround 360 video camera.
\newblock 2016.

\end{thebibliography}
}

\end{document}